\documentclass{article}

\usepackage[final]{neurips_2023}

\usepackage[utf8]{inputenc} 
\usepackage[T1]{fontenc}    
\usepackage{hyperref}       
\usepackage{url}            
\usepackage{booktabs}       
\usepackage{amsfonts}       
\usepackage{nicefrac}       
\usepackage{microtype}      
\usepackage{xcolor}         
\usepackage{multirow}
\usepackage{array}
\usepackage{graphicx}
\usepackage{longtable}
\usepackage{ctable}
\usepackage{enumitem}
\usepackage[normalem]{ulem}
\useunder{\uline}{\ul}{}

\title{On the Performance of Multimodal Language Models}

\author{%
  Utsav Garg \\
  Scale AI \\
  \texttt{utsav.garg@scale.com} \And
  Erhan Bas\thanks{work done while Erhan Bas was at Scale AI.} \\
  GE HealthCare \\
  \texttt{erhan.bas@gehealthcare.com} \\
}

\begin{document}

\maketitle

\begin{abstract}
Instruction-tuned large language models (LLMs) have demonstrated promising zero-shot generalization capabilities across various downstream tasks. Recent research has introduced multimodal capabilities to LLMs by integrating independently pretrained vision encoders through model grafting. These multimodal variants undergo instruction tuning, similar to LLMs, enabling effective zero-shot generalization for multimodal tasks. This study conducts a comparative analysis of different multimodal instruction tuning approaches and evaluates their performance across a range of tasks, including complex reasoning, conversation, image captioning, multiple-choice questions (MCQs), and binary classification. Through rigorous benchmarking and ablation experiments, we reveal key insights for guiding architectural choices when incorporating multimodal capabilities into LLMs. However, current approaches have limitations; they do not sufficiently address the need for a diverse multimodal instruction dataset, which is crucial for enhancing task generalization. Additionally, they overlook issues related to truthfulness and factuality when generating responses. These findings illuminate current methodological constraints in adapting language models for image comprehension and provide valuable guidance for researchers and practitioners seeking to harness multimodal versions of LLMs.
\end{abstract}

\section{Introduction}

Large instruction-tuned language models (LLMs) have emerged as powerful models showcasing remarkable zero-shot generalization capabilities across a diverse spectrum of downstream tasks. By learning to interpret natural language instructions, these models obviate the need for task-specific training. However, real-world applications, often involve multimodal data, such as images and text, necessitating the combination of visual and linguistic information for accurate and robust inference. To address this challenge, recent research~\cite{blip2, ye2023mplug, minigpt, llava} has introduced multimodal variants of LLMs, which integrate independently pretrained large vision encoders with LLMs. These models further undergo instruction tuning, aiming to leverage the combined power of visual and linguistic understanding.

In this paper, we conduct a comprehensive investigation that centers on comparing various approaches to multimodal instruction tuning and assessing their performance on a wide array of downstream tasks. Our study seeks to illuminate the efficacy, generalization capabilities, and limitations of publicly available models and their ablations across various domains. We evaluate the different approaches across a diverse range of tasks from complex reasoning to captioning and classification, to test their generalization capabilities. Moreover, we aim to identify whether specific design choices prove more suitable for specific tasks.

For our experiments, we consider five publicly available approaches for multimodal adaptation of LLMs: BLIP-2~\cite{blip2}, InstructBLIP~\cite{instructblip}, LLaVA~\cite{llava}, MiniGPT4~\cite{minigpt} and mPLUG-Owl~\cite{ye2023mplug}. These approaches encompass a wide gamut of architectural choices for injecting LLMs with multimodal capabilities. More details about these respective models is in the appendix~\ref{compared}. By conducting thorough benchmarking and analysis, we aim to identify the strengths and weaknesses of different multimodal instruction tuning strategies. Furthermore, we conduct ablation experiments on previously untested combinations within these approaches to unearth the optimal method for integrating visual capabilities into a language model. This includes exploring the utility of the largest feasible frozen vision encoder, attaching a trained vision head to extract concise visual features, employing a linear mapping layer to project these into the linguistic domain, and fine-tuning the decoder on multimodal data samples.

\section{Datasets and Evaluation}

We systematically assess all approaches across two primary task categories to comprehensively gauge their capabilities across a broad spectrum of challenges:

\begin{enumerate}
    \item \textbf{Complex Reasoning:} In this task category, the model confronts questions that necessitate external knowledge and cannot be resolved solely based on the image and question. This test predominantly assesses the language model's performance and illuminates the extent to which multimodal instruction tuning influences the base LLM's regression.
    \item \textbf{Captioning, Classification, Conversation, MCQ:} These tasks become feasible for the LLM only through the incorporation of the visual encoder. The performance across these tasks serves as a robust indicator of the approach's generalization prowess, encompassing architectural choices and instruction tuning data.
\end{enumerate}

To evaluate the general Visual Question Answering (VQA) abilities, we use the test set curated by LLaVA~\cite{llava}. The dataset consists of 30 images from the COCO dataset, each associated with three distinct question types: \textit{conversation}, \textit{description}, and \textit{complex reasoning} types, resulting in a total of 90 instruction-answer pairs. The ground-truth answers for these questions are generated by GPT-4, which is given the captions and bounding boxes associated with the image for context. The evaluation involves ranking both the predicted and reference answers on a scale of 1 to 10 by GPT-4. The metric is the relative score of the predicted answer to GPT-4's answer. To address potential biases due to answer order, as demonstrated by~\cite{wang2023large}, we adopt the Balanced Position Calibration~\cite{wang2023large} (BPC) strategy proposed by them, averaging scores from both possible orderings. Additionally, to account for the stochasticity of the generation, we average scores across 5 generations by each approach. Despite these considerations to make the evaluation as robust as possible, there are still limitations as GPT-4 is not actually seeing the image and is instead using a proxy for visual information. Moreover, we cannot expect GPT-4 evaluations to be perfect underlining the current limitation of open-ended answer evaluations.

For captioning tasks, we evaluate on the \textit{val} set of the NoCaps~\cite{agrawal2019nocaps} dataset and report the CIDEr~\cite{vedantam2015cider} metric. For visual Multiple Choice Questions (MCQ), we evaluate on the \textit{test} split of the ScienceQA~\cite{lu2022learn} dataset. Here, we exclusively consider samples with available image context. For binary classification, we evaluate on the Visual Spatial Reasoning~\cite{Liu2022VisualSR} (VSR) dataset. We use the \textit{zero-shot test} split provided in the official GitHub repository.

For tasks that require generation, we use a beam size of 5. For MCQ, we measure the log-likelihood of generating each of the answer options.

\section{Experiments}

\subsection{Comparing Existing Approaches}

Table~\ref{tab:arch} highlights the main differences between existing approaches, include architectural choices and the data used for instruction tuning.

\begin{table}[h]
\renewcommand{\arraystretch}{1.3}
\resizebox{\textwidth}{!}{%
\begin{tabular}{lccccccc}
\hline
\multicolumn{1}{l|}{\multirow{2}{*}{\textbf{Approach}}} & \multicolumn{1}{c|}{\multirow{2}{*}{\textbf{Instruction Data Size}}} & \multicolumn{2}{c|}{\textbf{Vision Encoder}}                                  & \multicolumn{2}{c|}{\textbf{Vision Head}}                                 & \multicolumn{2}{c}{\textbf{LLM}}                     \\ \cline{3-8} 
\multicolumn{1}{l|}{}                                   & \multicolumn{1}{c|}{}                                                & \multicolumn{1}{c|}{\textbf{Backbone}} & \multicolumn{1}{c|}{\textbf{Frozen}} & \multicolumn{1}{c|}{\textbf{Type}} & \multicolumn{1}{c|}{\textbf{Frozen}} & \multicolumn{1}{c|}{\textbf{Type}} & \textbf{Frozen} \\ \hline
BLIP-2                                                  & N/A                                                                  & ViT-g                                  & Yes                                  & Q-Former                           & No                                   & Vicuna-7B                          & Yes             \\
MiniGPT-4                                               & 3.5K                                                                 & ViT-g                                  & Yes                                  & Q-Former                           & Yes                                  & Vicuna-7B                          & Yes             \\
LLaVA                                                   & 150K                                                                 & ViT-L                                  & Yes                                  & N/A                                & N/A                                  & Vicuna-7B                          & No              \\
mPLUG-Owl                                               & 290K                                                                 & ViT-L                                  & No                                   & Similar to Perceiver Resampler     & No                                   & LLaMA-7B                           & No              \\
InstructBLIP                                           & 15M                                                                  & ViT-g                                  & Yes                                  & Q-Former                           & No                                   & Vicuna-7B                          & Yes             \\ \hline
\end{tabular}%
}
\vspace{0.1cm}
\caption{Details of the major components of the compared approaches. We compare variants of each approach with a similar decoder size. The instruction data sizes are approximate.}
\label{tab:arch}
\end{table}

For fair comparison, we compare variants of all approaches with a similarly sized LLM. Two of the approaches (LLaVA and mPLUG-Owl) use a smaller vision encoder (ViT-L), while the others use ViT-g. BLIP-2 does not do any multimodal instruction tuning, while all other approaches are instruction tuned on varying amounts of multimodal data. The vision encoder is frozen in all approaches except mPLUG-Owl, while the vision head is always trained when used, except for MiniGPT-4. Additionally, both of the approaches using the smaller vision encoder fine-tune their LLMs, while the others keep the language model frozen.

\subsection{Performance on Benchmarks}

Table~\ref{tab:results} presents the results on the various benchmarks discussed in Section 2. 

\begin{table}[h]
\centering
\renewcommand{\arraystretch}{1.3}
\resizebox{\textwidth}{!}{%
\begin{tabular}{lccccccc}
\hline
\multicolumn{1}{l|}{\multirow{2}{*}{\textbf{Approach}}} & \multicolumn{4}{c|}{\textbf{LLaVA VQA}} & \multicolumn{1}{c|}{\multirow{2}{*}{\textbf{VSR}}} & \multicolumn{1}{c|}{\multirow{2}{*}{\textbf{\begin{tabular}[c]{@{}c@{}}ScienceQA\\  (Image)\end{tabular}}}} & \multirow{2}{*}{\textbf{NoCaps}} \\ \cline{2-5}
\multicolumn{1}{l|}{} & \multicolumn{1}{c|}{\textbf{Conversation}} & \multicolumn{1}{c|}{\textbf{Detail}} & \multicolumn{1}{c|}{\textbf{Complex Reasoning}} & \multicolumn{1}{c|}{\textbf{Overall}} & \multicolumn{1}{c|}{} & \multicolumn{1}{c|}{} &  \\ \hline
BLIP-2 & 61.5 & 66.7 & 53.3 & 60.5 & 50 & {\ul 53.8} & {\ul 107.5} \\
MiniGPT-4 & 72.5 & 68.2 & 76.4 & 72.4 & {\ul 57.2} & 36.04 & 86.91 \\
LLaVA & {\ul 75.1} & {\ul 70.6} & {\ul 88.2} & {\ul 78} & 52.45 & 34.8 & 67.75 \\
mPLUG-Owl & 74 & 69.8 & 86.8 & 76.9 & 54.99 & 34.01 & 70.82 \\
InstructBLIP & \textbf{85.3} & \textbf{75.8} & \textbf{88.6} & \textbf{83.3} & \textbf{58.51} & \textbf{59.49} & \textbf{123.65} \\ \hline
\end{tabular}%
}
\vspace{0.1cm}
\caption{Comparison of all publicly available multimodal variants of LLMs on the four datasets discussed above. The scores for LLaVA VQA are relative scores compared to the reference answer as determined by GPT-4. For NoCaps, we use CIDEr and report accuracy for VSR and ScienceQA. The best results for each dataset are in \textbf{bold}, and the second-best results are {\ul underlined}.}
\label{tab:results}
\end{table}

For mPLUG-Owl, we use the checkpoint provided by the authors, where the full decoder is fine-tuned instead of the one that uses LORA~\cite{hu2021lora}, as it performs better overall. We ran all benchmarks using the checkpoints provided by the official implementations but directly used the results provided for BLIP-2 in the InstructBLIP paper for VSR, ScienceQA, and NoCaps.

The results demonstrate that InstructBLIP, which performs multitask instruction tuning on a variety of datasets, performs the best overall on all tasks, highlighting the importance of data diversity during instruction tuning. In contrast, all other instruction-tuned approaches perform poorly on tasks they were not trained on, likely due to overfitting to a specific task type. This is evident from the results of LLaVA, mPLUG-Owl, and InstructBLIP, all of which were trained on the LLaVA-150K data and are thus the top three performers on LLaVA VQA. However, only InstructBLIP continues to perform well on out-of-distribution tasks due to its richer instruction tuning data. There are qualitative samples from each of the three LLaVA benchmark categories in the appendix~\ref{fig:compare_qual}.

\subsection{Ablations on Model Components}

In this section, we conduct ablations on the LLaVA architecture to analyze the effects and importance of changes to different components, namely the vision encoder, vision head, and data. We chose the LLaVA architecture as the base because its training/evaluation code and data are available, and the LLaVA-150K training set has been used by other approaches as well.

\subsubsection{Vision Head}

The vision head operates over the patch embeddings of a frozen vision encoder (e.g., CLIP) and compresses the features to extract relevant details. In this section, we specifically compare the effects of using a Querying Transformer (Q-Former) as the vision head. The Q-Former is a small Transformer model that learns queries to extract relevant visual features from the vision encoder. For all experiments in this section (unless otherwise mentioned), we use an 80K balanced subset of the LLaVA data provided by the authors in the official GitHub repository.

\begin{table}[h]
\centering
\resizebox{\textwidth}{!}{%
\renewcommand{\arraystretch}{1.3}
\begin{tabular}{ccccccccc}
\hline
\multicolumn{2}{c|}{\textbf{Vision Head}} & \multicolumn{4}{c|}{\textbf{LLaVA VQA}} & \multicolumn{1}{c|}{\multirow{2}{*}{\textbf{VSR}}} & \multicolumn{1}{c|}{\multirow{2}{*}{\textbf{\begin{tabular}[c]{@{}c@{}}ScienceQA\\  (Image)\end{tabular}}}} & \multirow{2}{*}{\textbf{NoCaps}} \\ \cline{1-6}
\multicolumn{1}{c|}{\textbf{Type}} & \multicolumn{1}{c|}{\textbf{Frozen}} & \multicolumn{1}{c|}{\textbf{Conversation}} & \multicolumn{1}{c|}{\textbf{Detail}} & \multicolumn{1}{c|}{\textbf{Complex Reasoning}} & \multicolumn{1}{c|}{\textbf{Overall}} & \multicolumn{1}{c|}{} & \multicolumn{1}{c|}{} &  \\ \hline
N/A & N/A & 74.8 & 72.5 & 88.2 & 78.5 & 51.88 & 34.95 & 61.25 \\
Q-Former & Yes & 75 & 70.6 & 87.4 & 77.6 & 51.47 & \textbf{35.35} & 71.99 \\
Q-Former & No & \textbf{75.5} & \textbf{75.1} & \textbf{89.9} & \textbf{80.2} & \textbf{52.7} & \textbf{35.35} & \textbf{75.45} \\ \hline
\end{tabular}%
}
\vspace{0.1cm}
\caption{Comparison of the effect of having and training a vision head. All configurations use the same frozen vision backbone and trained LLM.}
\label{tab:com_vh}
\end{table}

The results in Table~\ref{tab:com_vh} clearly show that training the vision head over a frozen encoder offers a significant improvement compared to not having a vision head. However, using a frozen vision head does not provide any additional benefit compared to not having one. The improvement in performance comes from training the vision head along with the decoder to align features based on the task requirements.

\subsubsection{Variations of Vision Head}

In the previous section, we compared the effect of having a vision head. However, it is also possible to have different types of vision heads. The Q-Former architecture supports attending to both text and image together. For instruction tuning, it can be beneficial to have the Q-Former attend to the instruction along with the image, learning image features that are relevant for answering the question.

\begin{table}[h]
\vspace{0.1cm}
\centering
\resizebox{\textwidth}{!}{%
\renewcommand{\arraystretch}{1.3}
\begin{tabular}{ccccccccc}
\hline
\multicolumn{2}{c|}{\textbf{Vision Head}} & \multicolumn{4}{c|}{\textbf{LLaVA VQA}} & \multicolumn{1}{c|}{\multirow{2}{*}{\textbf{VSR}}} & \multicolumn{1}{c|}{\multirow{2}{*}{\textbf{\begin{tabular}[c]{@{}c@{}}ScienceQA\\  (Image)\end{tabular}}}} & \multirow{2}{*}{\textbf{NoCaps}} \\ \cline{1-6}
\multicolumn{1}{c|}{\textbf{Type}} & \multicolumn{1}{c|}{\textbf{Frozen}} & \multicolumn{1}{c|}{\textbf{Conversation}} & \multicolumn{1}{c|}{\textbf{Detail}} & \multicolumn{1}{c|}{\textbf{Complex Reasoning}} & \multicolumn{1}{c|}{\textbf{Overall}} & \multicolumn{1}{c|}{} & \multicolumn{1}{c|}{} &  \\ \hline
Image Q-Former & No & 75.5 & \textbf{75.1} & \textbf{89.9} & \textbf{80.2} & 52.7 & 35.35 & \textbf{75.45} \\
MultiModal Q-Former & No & 72.7 & 74 & 87 & 77.8 & 50.98 & 35.6 & 47.74 \\
MultiModal Q-Former$^*$ & No & \textbf{77.5} & 73 & 88.3 & 79.6 & \textbf{54.17} & \textbf{36.29} & 48.2 \\ \hline
\end{tabular}%
}
\vspace{0.1cm}
\caption{Comparison of the effect of having an image-only vs multimodal vision head. All configurations use the same frozen vision backbone and trained LLM.}
\label{tab:com_vh2}
\end{table}

Table~\ref{tab:com_vh2} presents the results of comparing different types of vision heads. In the "Image Q-Former" row, the Q-Former attends only to the image, while in the "MultiModal Q-Former" row, the Q-Former attends to both the instruction and the image. However, as the instruction tuning data contains multi-turn conversations, the instructions for all turns are concatenated together, which leads to suboptimal performance. The "MultiModal Q-Former$^*$" row modifies the instruction data by breaking all multi-turn conversation samples into separate training points. This version performs mostly on par with the image-only version. The drop in performance is observed in the detail category and captioning (NoCaps), as we do not want to restrict the features to the instruction in these cases, and it is best for the decoder to receive all possible information about the image. These experiments indicate that using instruction-aware visual features does not offer any advantage with open-ended question answering. However, where the final objective is classification, it offers some advantage as also observed by ~\cite{instructblip}.

\subsubsection{Size of Vision Encoder}

In the previous experiments, we used ViT-L as the frozen vision encoder. Here, we compare the effect of using a larger vision encoder, specifically ViT-g, for the different configurations mentioned above.

\begin{table}[h]
\vspace{0.1cm}
\centering
\resizebox{\textwidth}{!}{%
\renewcommand{\arraystretch}{1.3}
\begin{tabular}{cccccccccc}
\hline
\multicolumn{3}{c|}{\textbf{Vision Encoder}} & \multicolumn{4}{c|}{\textbf{LLaVA VQA}} & \multicolumn{1}{c|}{\multirow{2}{*}{\textbf{VSR}}} & \multicolumn{1}{c|}{\multirow{2}{*}{\textbf{\begin{tabular}[c]{@{}c@{}}ScienceQA\\  (Image)\end{tabular}}}} & \multirow{2}{*}{\textbf{NoCaps}} \\ \cline{1-7}
\multicolumn{1}{c|}{\textbf{Backbone}} & \multicolumn{1}{c|}{\textbf{$\#$ Params}} &  \multicolumn{1}{c|}{\textbf{Frozen}} & \multicolumn{1}{c|}{\textbf{Conversation}} & \multicolumn{1}{c|}{\textbf{Detail}} & \multicolumn{1}{c|}{\textbf{Complex Reasoning}} & \multicolumn{1}{c|}{\textbf{Overall}} & \multicolumn{1}{c|}{} & \multicolumn{1}{c|}{} &  \\ \hline
ViT-L & 304M & Yes & 75.5 & 75.1 & 89.9 & 80.2 & \textbf{52.7} & 35.35 & 75.45 \\
ViT-g & 1.0B & Yes & \textbf{81.5} & \textbf{79.1} & \textbf{90.9} & \textbf{83.9} & 50.16 & \textbf{35.6} & \textbf{90.89} \\ \hline
\end{tabular}%
}
\vspace{0.1cm}
\caption{Comparison of the effect of using a larger frozen vision encoder. All configurations use a trained image-only Q-Former and trained LLM.}
\label{tab:com_ve}
\end{table}

Table~\ref{tab:com_ve} presents the results comparing the ViT-L and ViT-g vision encoders, with all other configurations remaining the same. The results clearly demonstrate that having a larger vision encoder (ViT-g) helps improve performance across the board, even on tasks for which the model was not specifically trained (NoCaps). The larger encoder produces a richer image representation, leading to better overall performance.

\subsubsection{Effect of Data Size in Alignment Phase}

In all the discussed approaches, an alignment stage precedes instruction tuning, where a projection layer is learned to map the outputs of the vision heads to the input space of the language model. In this section, we study the effect of data size during the alignment phase. We use BLIP-2 as the starting point, which has been aligned on 129M image-text pairs. For comparison, we perform alignment using 595K image-text pairs, as done in LLaVA, for the same architecture.

\begin{table}[h]
\vspace{0.1cm}
\centering
\resizebox{\textwidth}{!}{%
\renewcommand{\arraystretch}{1.3}
\begin{tabular}{cccccccc}
\hline
\multicolumn{1}{c|}{\multirow{2}{*}{\textbf{Amount of Alignment Data}}} & \multicolumn{4}{c|}{\textbf{LLaVA VQA}} & \multicolumn{1}{c|}{\multirow{2}{*}{\textbf{VSR}}} & \multicolumn{1}{c|}{\multirow{2}{*}{\textbf{\begin{tabular}[c]{@{}c@{}}ScienceQA\\  (Image)\end{tabular}}}} & \multirow{2}{*}{\textbf{NoCaps}} \\ \cline{2-5}
\multicolumn{1}{c|}{} & \multicolumn{1}{c|}{\textbf{Conversation}} & \multicolumn{1}{c|}{\textbf{Detail}} & \multicolumn{1}{c|}{\textbf{Complex Reasoning}} & \multicolumn{1}{c|}{\textbf{Overall}} & \multicolumn{1}{c|}{} & \multicolumn{1}{c|}{} &  \\ \hline
129M & \textbf{82.6} & 76.5 & 90 & 83.1 & \textbf{52.21} & \textbf{35.65} & \textbf{91.07} \\
595K & 81.5 & \textbf{79.1} & \textbf{90.9} & \textbf{83.9} & 50.16 & 35.6 & 90.89 \\ \hline
\end{tabular}%
}
\vspace{0.1cm}
\caption{Comparison of the effect of data size during the alignment phase. All configurations use a trained image-only Q-Former and trained LLM.}
\label{tab:com_data}
\end{table}

Table~\ref{tab:com_data} presents the results comparing the alignment with different amounts of data. We observe that aligning the model with a larger amount of data does not significantly improve in-distribution downstream performance when the model is further instruction tuned. This is reasonable as the projection layer, which is trained during alignment, has a small number of parameters. Therefore, increasing the alignment data size eventually leads to diminishing returns. However, for out-of-distribution tasks (VSR, ScienceQA, NoCaps), the larger amount of alignment data does lead to small improvement in performance, most likely due to more robust features.

\begin{table}[h]
\vspace{0.1cm}
\centering
\resizebox{\textwidth}{!}{%
\renewcommand{\arraystretch}{1.3}
\begin{tabular}{cccccccc}
\hline
\multicolumn{1}{c|}{\multirow{2}{*}{\textbf{Amount of LLaVA VQA Data}}} & \multicolumn{4}{c|}{\textbf{LLaVA VQA}} & \multicolumn{1}{c|}{\multirow{2}{*}{\textbf{VSR}}} & \multicolumn{1}{c|}{\multirow{2}{*}{\textbf{\begin{tabular}[c]{@{}c@{}}ScienceQA\\  (Image)\end{tabular}}}} & \multirow{2}{*}{\textbf{NoCaps}} \\ \cline{2-5}
\multicolumn{1}{c|}{} & \multicolumn{1}{c|}{\textbf{Conversation}} & \multicolumn{1}{c|}{\textbf{Detail}} & \multicolumn{1}{c|}{\textbf{Complex Reasoning}} & \multicolumn{1}{c|}{\textbf{Overall}} & \multicolumn{1}{c|}{} & \multicolumn{1}{c|}{} &  \\ \hline
150K & \textbf{75.1} & 70.6 & \textbf{88.2} & 78 & \textbf{52.45} & 34.8 & \textbf{67.75} \\
80K & 74.8 & \textbf{72.5} & \textbf{88.2} & \textbf{78.5} & 51.88 & \textbf{34.95} & 61.25 \\ \hline
\end{tabular}%
}
\vspace{0.1cm}
\caption{Comparison of the effect of data size during the instruction tuning phase. All configurations use a ViT-L vision encoder, no vision head, and a trained LLM.}
\label{tab:stage2_data}
\end{table}

Table~\ref{tab:stage2_data} shows a similar trend for the quantity of data used in instruction tuning. In all the ablation experiments, we used an 80K balanced subset of the LLaVA instruction data. However, if we compare the performance of our recreated versions of LLaVA in the table above, on the original 150K and the 80K data subset, they perform at par. Beyond a certain point, increasing the amount of data does not significantly improve performance. However, data diversity does play a role, as demonstrated by the superior performance of InstructBLIP in Table~\ref{tab:results}.

\subsubsection{Training the Language Model}

The ablations discussed so far all fine-tune the language model following LLaVA. This section compares the effect of training or keeping the decoder frozen across various configurations.

\begin{table}[h]
\vspace{0.1cm}
\resizebox{\textwidth}{!}{%
\renewcommand{\arraystretch}{1.3}
\begin{tabular}{ccccccccccc}
\hline
\multicolumn{1}{c|}{\multirow{2}{*}{\textbf{Vision Encoder}}} & \multicolumn{2}{c|}{\textbf{Vision Head}} & \multicolumn{1}{c|}{\multirow{2}{*}{\textbf{LLM Frozen}}} & \multicolumn{4}{c|}{\textbf{LLaVA VQA}} & \multicolumn{1}{c|}{\multirow{2}{*}{\textbf{VSR}}} & \multicolumn{1}{c|}{\multirow{2}{*}{\textbf{\begin{tabular}[c]{@{}c@{}}ScienceQA\\  (Image)\end{tabular}}}} & \multirow{2}{*}{\textbf{NoCaps}} \\ \cline{2-3} \cline{5-8}
\multicolumn{1}{c|}{} & \multicolumn{1}{c|}{\textbf{Type}} & \multicolumn{1}{c|}{\textbf{Frozen}} & \multicolumn{1}{c|}{} & \multicolumn{1}{c|}{\textbf{Conversation}} & \multicolumn{1}{c|}{\textbf{Detail}} & \multicolumn{1}{c|}{\textbf{Complex Reasoning}} & \multicolumn{1}{c|}{\textbf{Overall}} & \multicolumn{1}{c|}{} & \multicolumn{1}{c|}{} &  \\ \hline
ViT-L & N/A & N/A & Yes & 58.1 & 66.2 & 75.8 & 66.7 & 50.65 & 34.9 & 54.31 \\
ViT-L & N/A & N/A & No & \textbf{74.8} & \textbf{72.5} & \textbf{88.2} & \textbf{78.5} & \textbf{51.88} & \textbf{34.95} & \textbf{61.25} \\ \hline
ViT-L & Q-Former & No & Yes & 68.4 & 69.1 & 85 & 74.1 & 49.75 & \textbf{35.45} & 67.18 \\
ViT-L & Q-Former & No & No & \textbf{75.5} & \textbf{75.1} & \textbf{89.9} & \textbf{80.2} & \textbf{52.7} & 35.35 & \textbf{75.45} \\ \hline
ViT-g & Q-Former & No & Yes & 72.3 & 77.6 & 88 & 79.3 & \textbf{50.49} & \textbf{36.49} & 89.98 \\
Vit-g & Q-Former & No & No & \textbf{81.5} & \textbf{79.1} & \textbf{90.9} & \textbf{83.9} & 50.16 & 35.6 & \textbf{90.89} \\ \hline
\end{tabular}%
}
\vspace{0.1cm}
\caption{Comparing the effect of training the decoder. The configurations mention the vision encoder and head used. All experiments use the same LLM (Vicuna-7B).}
\label{tab:com_dec}
\end{table}

The experiments in Table~\ref{tab:com_dec} show that when instruction tuning on the LLaVA dataset, training the decoder helps in almost all scenarios. The performance improvements are reduced to an extent when using a larger vision backbone (ViT-g), but there is still some gain. InstructBLIP uses a frozen decoder, but these experiments suggest that we can achieve an additional boost in performance if the decoder is trained as well. Moreover, as the gradients are always propagated through the decoder in this architecture setup, training the decoder does not incur any significant overhead.

\section{Discussion and Conclusion}

The benchmarks and ablation experiments offer various insights into multimodal LLMs. The main takeaways from this study are as follows:

\begin{itemize}
    \item \textbf{Vision Encoder:} Using a larger vision encoder (ViT-L vs. ViT-g) consistently improves performance across all tasks, as it captures a richer image representation. However, fine-tuning the vision encoder does not improve performance on downstream tasks as observed from the relative performance of mPLUG-Owl in Table~\ref{tab:results}, which does not keep it frozen.
    \item \textbf{Vision Head:} Fine-tuning a vision head (e.g., Q-Former) is powerful as it enables the model to extract a better representation for the downstream task while also speeding up training and inference due to the smaller output representation. However, using a multimodal vision head that encodes both the image and instruction together does not show any apparent advantage over using an image-only head in open-ended question answering. This suggests that it is more important to condense all visual information and pass that to the language model through grafting and letting the LLM use the full context to answer questions.
    \item \textbf{LLM:} Training the Large Language Model (LLM) during instruction tuning can lead to additional performance gains without significant overhead in training cost, as gradients are propagated through the LLM in all cases. If there are concerns of LLM regression on text only tasks, we can use adapters (LoRA~\cite{hu2021lora}/IA3~\cite{liu2022few} etc.) to leave the base model unaltered and use the adapter version when inputs are multimodal.
    \item \textbf{Data:} The size of the training dataset becomes less important beyond a certain point in both the alignment and instruction tuning stages. However, data diversity, both within a task and across tasks, remains crucial for achieving superior performance as shown by results of InstructBLIP.
\end{itemize}

These results provide valuable insights for future research in this direction, guiding architecture choices and emphasizing the importance of data diversity both within and across tasks. The findings suggest that focusing on larger frozen vision encoders, training vision heads, and optimizing the LLM can yield improvements in performance on multimodal tasks. Moreover, exploring diverse and representative datasets can contribute to achieving state-of-the-art performance.

This work also highlights that differences in architectural methodologies for grafting multimodal capabilities into LLMs. A simple strategy of extracting relevant condensed visual features and transforming them via linear projections to the language space performs as well as any other. The major areas to focus on are data curation and task diversity. Another challenge the community should focus on are hallucinations of these multimodal variants. Language models are unaware of visual concepts during pretraining and tend to hallucinate objects/concepts in images that might not exist when probed about them instead of answering no. To make these models useful, a focus on alleviating these hallucinations will be essential.

In conclusion, this study sheds light on the key components and strategies for building effective multimodal Large Language Models. It focuses on the limitations of current research, and highlights focus areas that will bring the largest impact to their capabilities and usefulness.

\newpage

\bibliographystyle{unsrtnat}
\bibliography{mybib}

\appendix
\section{Appendix}

\subsection{Compared Approaches}
\label{compared}
For our experiments, we consider the five publicly available approaches listed below. 

\begin{itemize}
    \item \textbf{BLIP-2~\cite{blip2}:} Aligns a frozen vision encoder and frozen language model through a Querying Transformer (Q-Former) and a linear projection layer. The Q-Former is a small Transformer model that learns queries to extract relevant visual features from the vision encoder. The model is trained in two stages: first, to learn the image representation, and then to translate the learned representation through the LLM. It is pretrained with 129M image-text pairs from COCO~\cite{lin2014microsoft}, Visual Genome~\cite{krishna2017visual}, CC3M~\cite{sharma2018conceptual}, CC12M~\cite{changpinyo2021conceptual}, SBU~\cite{ordonez2011im2text}, and LAION400M~\cite{schuhmann2021laion}.
    \item \textbf{InstructBLIP~\cite{instructblip}:} Uses the same architecture as BLIP-2 and performs multimodal instruction tuning. During instruction tuning, the instruction is also passed along with the image to the Q-Former to extract instruction-relevant visual features. It translates multiple captioning and VQA datasets into an instruction-answer format and performs instruction tuning in a multi-task setup. It is trained on around 15M image-instruction pairs from COCO Caption~\cite{lin2014microsoft}, Web CapFilt~\cite{li2022blip}, TextCaps~\cite{sidorov2020textcaps}, VQAv2~\cite{goyal2017making}, OKVQA~\cite{okvqa}, A-OKVQA~\cite{AOKVQA}, OCR-VQA~\cite{mishraICDAR19}, and LLaVA-150K~\cite{llava}.
    \item \textbf{LLaVA~\cite{llava}:} Has a CLIP~\cite{radford2021learning} ViT-L vision encoder mapped to a Vicuna decoder through a linear layer. It follows a two-stage training process: in stage-1 (alignment), only the linear mapper is trained to align the visual output with the language input. In stage-2, both the mapper and LLM are trained to output answers to image-instruction pairs. For stage-1, 595K image-text pairs from CC3M~\cite{sharma2018conceptual} are used. For stage-2, they create a dataset of 158K image-instruction pairs across conversation, description, and complex reasoning tasks by prompting the language-only GPT-4~\cite{OpenAI2023GPT4TR}.
    \item \textbf{MiniGPT4~\cite{minigpt}:} Follows the same architecture as BLIP-2 but uses only the pretrained Q-Former and learns the linear mapping layer through the same alignment process as LLaVA. It also follows a two-stage training paradigm, but in both stages, only the linear mapping layer is trained. For stage-1, they use 5M image-text pairs from LAION, SBU, and Conceptual Captions. For stage-2, they self-curate a high-quality instruction dataset with 3.5K instruction-image pairs to improve the model's conversational ability.
    \item \textbf{mPLUG-Owl~\cite{ye2023mplug}:} This approach uses a CLIP ViT-L vision encoder, a visual abstractor which follows a similar architecture to Perceiver Resampler~\cite{alayrac2022flamingo}, a linear mapping layer, and a LLaMA~\cite{touvron2023llama} decoder. They also follow a two-stage training setup, but in this work, all parts of the model are trained during one of the two stages. The visual abstractor serves a similar purpose to the Q-Former and is trained in stage-1 along with the visual encoder to extract relevant features. In stage-2, the visual encoder and abstractor are frozen, and the LLM is trained. For stage-1, they use a massive dataset of about 1.1B image-text pairs from LAION-400M, COYO-700M~\cite{kakaobrain2022coyo-700m}, Conceptual Captions, and MSCOCO. For stage-2, they use around 290K instruction samples from Alpaca~\cite{alpaca}, Vicuna~\cite{vicuna2023}, Baize~\cite{xu2023baize}, and LLaVA-150K.
\end{itemize}

\newpage
\subsection{Qualitative comparison of benchmaked approaches}
\label{fig:compare_qual}

\renewcommand{\arraystretch}{1.5}
\begin{longtable}{m{4cm} | l | m{7cm}}
\specialrule{.1em}{.05em}{.05em} 
\multicolumn{1}{c|}{\textbf{Image and Question}} & \multicolumn{1}{c|}{\textbf{Model}} & \multicolumn{1}{c}{\textbf{Generations}} \\
\specialrule{.1em}{.05em}{.05em} 
\endhead
 
 \multirow{5}{*}{
 \begin{minipage}{.3\textwidth}
  \centering
  \vspace{0.4cm}
  \includegraphics[width=0.7\linewidth]{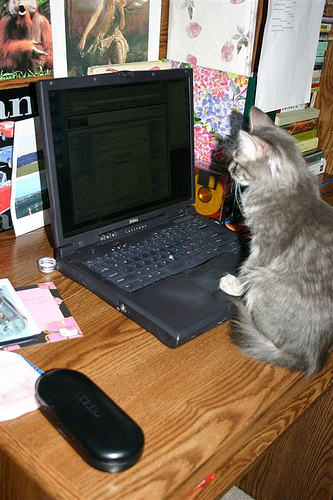}
  
  What is the main focus \\ of the image? \\
  \vspace{0.1cm}
  (Coversation)
 \end{minipage} 
 } 
 & BLIP-2 & Is it the cat or the laptop? \\ \cline{2-3} 
 & InstructBLIP & The main focus of the image is a cat sitting on a desk in front of a laptop computer. \\ \cline{2-3} 
 & LLaVA & The main focus of the image is a cat sitting on a desk and looking at a laptop. \\ \cline{2-3} 
 & MiniGPT-4 & The main focus of the image appears to be a cat sitting on top of a laptop computer that is open and connected to a keyboard. The cat is sitting on the right side of the laptop, facing the camera. The background of the image appears to be a cluttered desk with papers, pens, and other items scattered around. \\ \cline{2-3} 
 & mPLUG-Owl & The main focus of the image is on a cat sitting on top of a laptop and looking at the screen. \\ 
 \specialrule{.1em}{.05em}{.05em} 

 \multirow{5}{*}{
 \begin{minipage}{.3\textwidth}
  \centering
  \vspace{0.3cm}
  \includegraphics[width=0.75\linewidth]{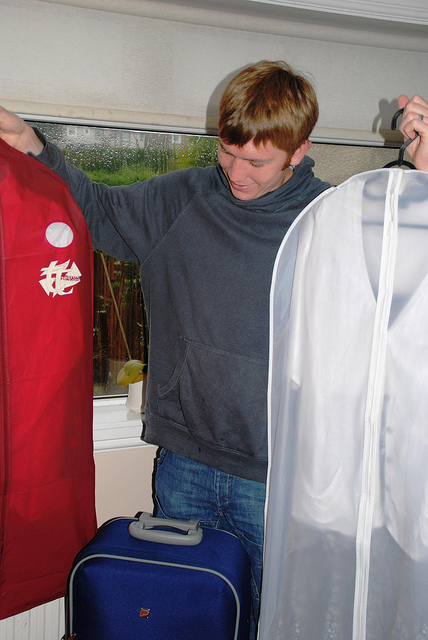}
  
  Why might the man be holding the clothes on hangers? \\
  \vspace{0.1cm}
  (Complex Reasoning)
 \end{minipage} 
 } 
 & BLIP-2 & He might be getting ready to pack for a trip. \\ \cline{2-3} 
 & InstructBLIP & The man might be holding the clothes on hangers because he is preparing to pack them into a suitcase. In the image, he is standing next to a window with a suitcase open, indicating that he is in the process of packing his belongings for a trip. By holding the clothes on hangers, he can ensure that they maintain their shape and don't get wrinkled or creased during the packing process. Additionally, hanging the clothes on hangers helps to conserve space in the suitcase and makes it easier to access and organize the items when unpacking at the destination. \\ \cline{2-3} 
 & LLaVA & The man could be holding the clothes on hangers for a variety of reasons. He might be preparing to hang the clothes in a closet or dressing area in his home, or he could be taking them to a dry cleaning store or laundromat to have them cleaned. The image does not provide enough context to determine the exact reason, but it is clear that the man is in the process of handling the clothes on hangers. \\ \cline{2-3} 
 & MiniGPT-4 & It is not clear why the man is holding the clothes on hangers. It could be that he is trying to decide what to wear, or he could be holding the clothes to take them to a dry cleaner. Without more information, it is difficult to say for certain why the man is holding the clothes on hangers. \\ \cline{2-3} 
 & mPLUG-Owl & The man could be holding the clothes on hangers as he is either organizing his closet, packing for a trip, or getting ready to hang them up in his room or a hotel room. Hanging the clothes in a well-organized manner can help keep them clean and prevent wrinkles, especially if he is packing for a trip. Additionally, he may be preparing for a photo shoot or attending an event where the clothing needs to be presented nicely. Holding the clothes on hangers helps him maintain their appearance and ensure they are ready for display, presentation, or use. \\ 
 \specialrule{.1em}{.05em}{.05em} 

 \multirow{5}{*}{
 \begin{minipage}{.3\textwidth}
  \centering
  \vspace{0.3cm}
  \includegraphics[width=0.8\linewidth]{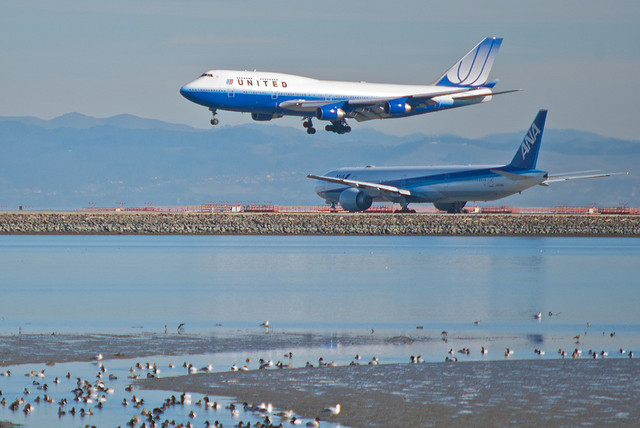}
  
  What do you see happening \\ in this image? \\
  \vspace{0.1cm}
  (Detail Description)
 \end{minipage} 
 } 
 & BLIP-2 & There are two airplanes in the picture. One is landing and the other is taking off. There are also a lot of birds in the water. The mountains are in the background. \\ \cline{2-3} 
 & InstructBLIP & The image depicts two large airplanes flying over a body of water, possibly an ocean or a lake. One of the planes is in the process of taking off, while the other is already airborne. There are several birds scattered throughout the scene, with some perched on the water's surface and others flying in the sky. In total, there are at least 10 birds visible in the image. The airplanes are positioned at different heights, with one closer to the water and the other higher up in the sky. Overall, the scene captures the beauty of aviation and nature coexisting in harmony. \\ \cline{2-3} 
 & LLaVA & The image features a large airplane taking off from an airport runway, with the airplane's wings fully extended. In the background, there is another airplane on the runway, possibly preparing for takeoff or landing. The scene is set on a body of water, with several birds in the vicinity of the airplanes. A flock of birds is resting near the first airplane, while another flock is situated in the background of the image. The combination of the airplanes, birds, and water contribute to a picturesque aerial view. \\ \cline{2-3} 
 & MiniGPT-4 & I see a large passenger jet airplane flying over a body of water with some birds in the foreground. \\ \cline{2-3} 
 & mPLUG-Owl & The image captures a large airplane taking off from a runway, surrounded by a group of birds. The birds appear to be flying in a nearby area, likely observing the airplane as it ascends. There is also a flock of birds flying in the air. \\ 
 \specialrule{.1em}{.05em}{.05em} 

\caption{Qualitative comparison of the publicly available multimodal LLMs on a selected sample of image-question pairs from the LLaVA test set.}
\label{table:qualitative}
\end{longtable}
\renewcommand{\arraystretch}{1}

\end{document}